\documentclass[letterpaper]{article} 
\usepackage{aaai25}  
\nocopyright
\usepackage{times}  
\usepackage{helvet}  
\usepackage{courier}  
\usepackage[hyphens]{url}  
\usepackage{graphicx} 
\usepackage{natbib}  
\usepackage{caption} 
\usepackage{algorithm}
\usepackage{algorithmic}
\usepackage{newfloat}
\usepackage{listings}
\DeclareCaptionStyle{ruled}{labelfont=normalfont,labelsep=colon,strut=off} 
\floatstyle{ruled}
\newfloat{listing}{tb}{lst}{}
\floatname{listing}{Listing}
\title{Navigating the Noisy Crowd: Finding Key Information for Claim Verification}

\author {
    Haisong Gong\textsuperscript{\rm 1,\rm 2},
    Huanhuan Ma\textsuperscript{\rm 1,\rm 2},
    Qiang Liu\textsuperscript{\rm 1,\rm 2},
    Shu Wu\textsuperscript{\rm 1,\rm 2},
    Liang Wang\textsuperscript{\rm 1,\rm 2}
}
\affiliations {
    \textsuperscript{\rm 1}New Laboratory of Pattern Recognition (NLPR)\\
Institute of Automation, Chinese Academy of Sciences\\
    \textsuperscript{\rm 2}School of Artificial Intelligence, University of Chinese Academy of Sciences\\
    \{gonghaisong2021, mahuanhuan2021\}@ia.ac.cn, \{qiang.liu, shu.wu, wangliang\}@nlpr.ia.ac.cn
}

\usepackage{bbding}
\usepackage{color}
\usepackage{amsfonts,amssymb,amsmath} 
\usepackage{xspace}
\usepackage{booktabs}
\usepackage{multirow}
\usepackage[many]{tcolorbox}    	
\newcommand{\themodel}{EACon\xspace}

\begin{document}

\maketitle

\begin{abstract}
Claim verification is a task that involves assessing the truthfulness of a given claim based on multiple evidence pieces. Using large language models (LLMs) for claim verification is a promising way. However, simply feeding all the evidence pieces to an LLM and asking if the claim is factual does not yield good results. The challenge lies in the noisy nature of both the evidence and the claim: evidence passages typically contain irrelevant information, with the key facts hidden within the context, while claims often convey multiple aspects simultaneously. To navigate this ``noisy crowd'' of information, we propose \themodel (Evidence Abstraction and Claim Deconstruction), a framework designed to find key information within evidence and verify each aspect of a claim separately.  \themodel first finds keywords from the claim and employs fuzzy matching to select relevant keywords for each raw evidence piece. These keywords serve as a guide to extract and summarize critical information into abstracted evidence. Subsequently, \themodel deconstructs the original claim into subclaims, which are then verified against both abstracted and raw evidence individually. We evaluate \themodel using two open-source LLMs on two challenging datasets.  Results demonstrate that \themodel consistently and substantially improve LLMs' performance in claim verification. 
\end{abstract}

\begin{figure*}[t]
\setlength{\abovecaptionskip}{0pt}  
\setlength{\belowcaptionskip}{0pt}  
   \begin{center}
   \includegraphics[width=1\textwidth]{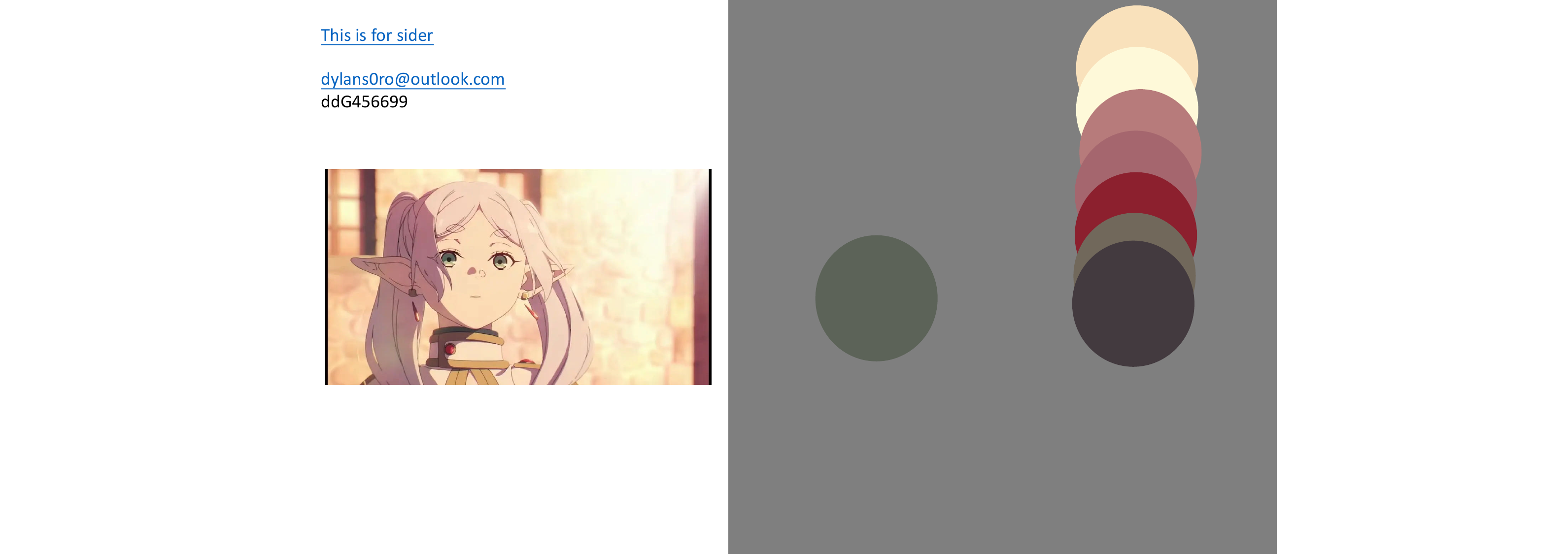}
   \end{center}
   
   \caption{Architecture of \themodel. The input is a claim and raw evidence, and the output is the predicted veracity of the claim. \themodel extracts keywords from the claim and uses fuzzy matching to select keywords for each piece of the raw evidence. These selected keywords are then used to summarize the raw evidence into abstracted evidence. \themodel then deconstructs the claim into subclaims, which are verified against both the raw and abstracted evidence using a zero-shot approach.} 
   \label{fig:model}
\end{figure*}

\section{Introduction}

The ease of creating and sharing information has led to a surge in misinformation within society, spanning from social media to prominent events like the U.S. Presidential debates, disrupting societal norms~\cite{bakir2018fake}. Consequently, the automated verification of information accuracy has become paramount. One critical aspect of this is claim verification, which involves using models to evaluate the truthfulness of a given statement (claim) based on multiple evidence pieces~\cite{guo2022survey}.

Claim verification can be viewed as a type of Natural Language Inference (NLI) task. Prior studies have delved into techniques such as fine-tuning pre-trained language models and utilizing graph neural networks to establish relationships between evidence in claim verification~\cite{ma2023ex,gong2024heterogeneous}. With recent advancements in large language models (LLMs)~\cite{zhao2023survey}, leveraging these models for claim verification holds significant promise.

Despite the potential of LLMs, applying them directly to claim verification by simply feeding all the evidence pieces and asking if a claim is factual falls short in yielding satisfactory outcomes. Even advanced methods, such as leveraging in-context examples through few-shot learning or enhancing LLM reasoning via strategies like Chain of Thought (CoT)~\cite{wei2022chain} or complex reasoning chains~\cite{fu2022complexity}, do not consistently improve claim verification outcomes~\cite{hu2023large}. 
This is because the task of claim verification necessitates not only reasoning abilities but also the capacity to handle the inherently ``noisy'' nature of evidence and claims, which both direct LLM applications and these prompt techniques struggle to address effectively.  

In the case of ``noisy'' evidence, an evidence piece may be rife with irrelevant information, while the key information occupies only a small portion and is hidden deeply within the context. This necessitates the model to possess the capability to sift through the noise and extract the pertinent information from the ``noisy evidence crowd.'' On the other hand, ``noisy'' claims are often expressed in a convoluted manner, encompassing multiple aspects simultaneously rather than presenting a concise, atomic statement. These ``noisy'' claims pose challenges for the direct application of LLMs. This is because LLMs typically tend to compare the overall semantic meaning between the evidence and claim, overlooking minor details. However, in the realm of claim verification, even minor inaccuracies should render a claim false, irrespective of the overall semantic coherence. 

To address this challenge, we propose the \textbf{\themodel} (Evidence Abstraction and Claim Deconstruction) framework. \themodel extracts and summarizes the key information from the raw evidence into abstracted evidence to aid LLM verification. It also deconstructs the claim into subclaims, allowing each aspect of the claim to be checked in detail, increasing the likelihood of identifying errors. In this framework, we design a keyword-based technique to extract keywords from the claim and use fuzzy matching to select relevant keywords as guidance to conduct evidence abstraction. This keyword-guided strategy mitigates the impact of conflicts between inaccurate claims and evidence content, while selecting relevant keywords by fuzzy matching aids in reducing the LLM's tendency to generate content not conveyed by evidence, as illustrated in Figure \ref{fig:explain}. Furthermore, for complex scenarios, we provide the LLM with contextual information about the original claim during the subclaim verification stage, further improving the model's performance.

In summary, our key contributions include: 
\begin{itemize} 
\item We highlight the key challenge in claim verification as navigating the ``noisy crowd'' of claim and evidence information, which hampers the performance of LLMs in claim verification.
\item We propose the \themodel framework, which extracts and summarizes the key information from raw evidence into abstracted evidence based on selected keywords and deconstructs the claim into subclaims for verification.
\item We demonstrate the effectiveness of \themodel on the HOVER and FEVEROUS-S datasets, using two open-source LLMs (Vicuna-13B and Mixtral-8x7B). The results show that \themodel can consistently and substantially improve LLMs' performance in claim verification. \end{itemize}

\section{Related Work}
\subsubsection{Claim Verification}
Traditional methods for claim verification can be categorized into two main approaches. The first approach employs pre-trained language models fine-tuned specifically for claim verification. These models either concatenate the evidence and claims into a single input~\cite{aly2021feverous,thorne2018fever,hu2022dual} or process each piece of evidence separately and then aggregate the results~\cite{soleimani2020bert, jiang2021exploring, gi-etal-2021-verdict}. The second approach utilizes graph neural networks to capture complex semantic interactions through evidence graphs~\cite{gi-etal-2021-verdict, Zhaotransxh, zhaoKgat, ZhongLCWQH20, Chenevidencenet, gong2024heterogeneous}.
Recent studies have explored leveraging the reasoning abilities of LLMs in verification tasks. For example, ProgramFC~\cite{pan2023fact} employs LLMs to generate reasoning programs that guide the verification process, while EX-FEVER~\cite{ma2023ex} elicits LLMs' capability to generate textual explanations for claim verification results. Factscore~\cite{min2023factscore} proposes fine-grained atomic evaluation for long text inputs. \citet{pan2023qacheck,chen2022generating,li2023self,rani2023factify} propose to generate a series of questions or queries for claim verification. However, none of these methods address the ``noisy'' problem of both evidence and claim information, which our method focuses on.

\subsubsection{Large Language Model Reasoning}
The reasoning capabilities of LLMs form the cornerstone for LLM-based verification tasks. In recent years, in-context learning, popularized by the few-shot prompting approach of \citet{Browngpt1}, has enabled models to generalize tasks from a few examples. The reasoning ability of LLMs has been further enhanced through various strategies, such as chain-of-thought prompting (CoT)~\cite{wei2022chain, wang2022self,kojima2022large}, which improves reasoning by generating intermediate steps in problem-solving. \citet{fu2022complexity} propose selecting complex reasoning examples as prompts to boost LLMs' reasoning performance. 
However, these strategies do not consistently enhance performance in claim verification tasks~\cite{hu2023large}.
Our method does not focus on enhancing the reasoning abilities of LLMs to solve claim verification tasks. Instead, it aims to improve claim verification performance by reducing ``noise'' through evidence abstraction and claim deconstruction, thereby leveraging the LLMs' strengths more effectively.


\newtcolorbox{mybox}{
    boxrule = 1.0pt,
    rounded corners,
    colback = sub,
    left = 1mm,
    right = 1mm,
    top = 1mm,
    bottom = 1mm,
    before skip = 1.6mm,
    arc = 5pt   
}
\definecolor{sub}{HTML}{ece8e8}

\begin{figure}[t]
\setlength{\abovecaptionskip}{0pt}  
\setlength{\belowcaptionskip}{0pt}  
   \begin{center}
   \includegraphics[width=1\columnwidth]{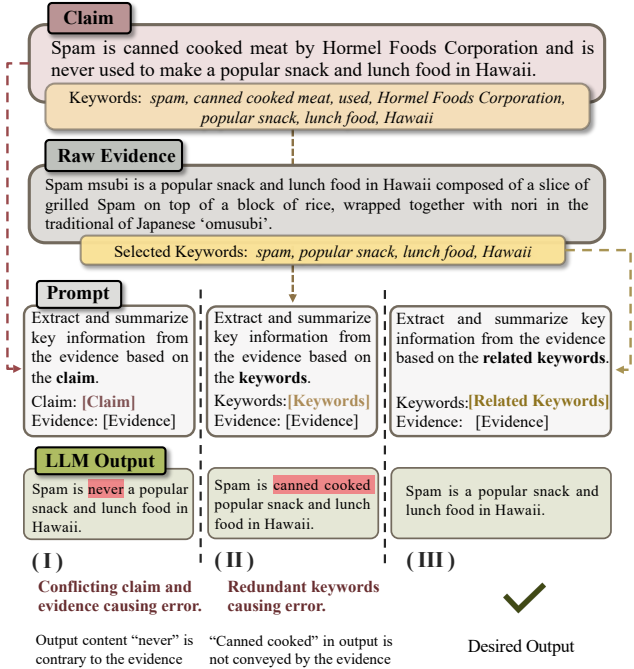}
   \end{center}
   \vspace{0cm}
   \caption{Illustration of different methods for prompting LLM to abstract evidence. From left to right: (I) Prompting LLM based on claim leads to incorrect output due to conflicting claim and evidence. (II) Prompting LLM with all keywords may result in generating content not supported by the evidence. (III) Our proposed method using selected keywords leads to correct output.} 
   \label{fig:explain}
\end{figure}


\section{Method}

In this section, we introduce the details of our proposed framework, \themodel. Generally, \themodel is composed of three major components: Evidence Abstraction, Claim Deconstruction, and Subclaim Verification. Both Evidence Abstraction and Claim Deconstruction are designed to address the ``noisy crowd'' problem for claim verification. After these two preparatory steps, the final Subclaim Verification component verifies each subclaim and produces the overall result. The architecture of \themodel is shown in Figure \ref{fig:model}.
\subsection{Task Formulation}
The objective of claim verification is to determine the veracity of a given claim based on multiple pieces of evidence. Typically, each piece of evidence is a sentence or paragraph drawn from sources like Wikipedia. Mathematically, given a claim $c$ and an evidence set containing $n$ piece of evidence $\mathcal{E} = \{e_1,e_2,\cdots,e_n\}$. the task is to find a model $\hat p = f(c, \mathcal{E})$ that outputs the predicted veracity $\hat p$, where $\hat p = \text{True}\ or\ \text{False}$.

\subsection{Evidence Abstraction}\label{section:EA}

Evidence Abstraction serves as the first part of \themodel. It involves processing each raw piece of evidence to extract useful information and eliminate noisy information. Instead of naively prompting an LLM to perform extraction and summarization tasks, we designed a keyword-based method.

As shown in Part I of Figure \ref{fig:explain}, if an LLM is prompted to summarize the evidence against a claim that is inherently false, there is a risk of conflicting information that could lead to subpar results. To alleviate this problem, we designed a keyword-based method. \themodel extracts keywords from the claim, which encapsulate the claim's essence without introducing any biases from the claim itself. This approach helps in circumventing potential conflicts. However, as shown in Part II of Figure \ref{fig:explain}, if all the keywords are used to summarize an evidence, the redundant keywords may lead the LLM to output content not conveyed by the evidence. To address this, we designed a Keyword Selection procedure to remove the irrelevant keywords, resulting in better summarization results. The effectiveness of the keyword-based design is empirically validated in Section \ref{section:WK}. Subsequent sections will delve into the details of the Evidence Abstraction process, encompassing Keyword Extraction, Keyword Selection, and Evidence Summarization.

\subsubsection{Keyword Extraction}
Keyword Extraction is the initial step in the Evidence Abstraction process, aimed at identifying essential keywords from the claim. These keywords, including important nouns, verbs, and phrases, serve as a guide for extracting information from evidence in subsequent sections. We instruct an LLM to perform keyword extraction and provide examples to aid its understanding of the task and output formatting. The process can be formally described as:
\begin{equation}
    \{k_1,k_2,\cdots,k_m\} = \arg \max p(\mathcal{K}|T_{KE},c;\theta_{LLM})
\end{equation}
where $k_i$ is the $i$th keyword selected by the LLM from the potential keywords set $\mathcal{K}$, and $m$ keywords are selected in total for the claim $c$. $\theta_{LLM}$ represents the LLM model, and $T_{KE}$ is the prompt template used for Keyword Extraction:

\begin{mybox}
\textbf{Task Description}: Extract key components such as important verbs, nouns, and phrases from the provided sentence. Focus on identifying and highlighting the most relevant elements.\\
\textbf{Instructions}:
Carefully read the input sentence.
Identify and list the significant verbs, nouns, and pertinent phrases.
Ensure the output succinctly encapsulates the essence of the input by focusing on these key components.\\
\textbf{Examples}:\\
Input: Spam is canned cooked meat by Hormel Foods Corporation is never used to make a popular snack and lunch food in Hawaii.\\
Output: spam, canned cooked meat, Hormel Foods Corporation, used, popular snack, lunch food, Hawaii.\\
\vspace{1.3mm}{{[}More Examples{]}}\\
Given the following input and keywords, provide a concise and factual summary based on the examples above. Exclude any information not directly related to the keywords.\\
\textbf{Input}: [Claim] ($c$)\\
\textbf{Output}:
\end{mybox}

\subsubsection{Keyword Selection}
We have obtained $m$ keywords from the claim $c$. Given the $i$th piece of evidence $e_i$, the second step is to identify which of these $m$ keywords are related to $e_i$. Since we do not want to use all the keywords as the evidence summary guidance, as described in the previous section and Figure \ref{fig:explain}, we employ fuzzy matching as a low-cost and efficient way to implement this task.

Fuzzy matching can be used to evaluate the similarity between a keyword and a piece of evidence. Specifically, we use two functions from the fuzzywuzzy package\footnote{https://github.com/seatgeek/thefuzz}, namely \texttt{partial\_ratio} and \texttt{token\_set\_ratio}. The \texttt{partial\_ratio} function computes the similarity (edit distance) of the best matching substring of the evidence to the input keyword, while the \texttt{token\_set\_ratio} function determines the similarity score of the intersection of unique tokens between the input keyword and the evidence piece. These functions are applied to compare each keyword $k_j, j\in \{1,2,\cdots,m\}$ with the $i$th piece of evidence $e_i$. To prevent the omission of potentially relevant keywords, those where either of the similarity scores exceeds a preset threshold will be selected. Mathematically:
\begin{multline}
    \mathcal{S}_i = \{k_j \mid \texttt{ partial\_ratio}(k_j, e_i) > t_1 \text{ or } \\ \texttt{ token\_set\_ratio}(k_j, e_i) > t_2, 1 \leq j \leq m\}
\end{multline}
where $\mathcal{S}_i$ indicates the selected keywords set for the $i$th piece of evidence $e_i$, $t_1$ and $t_2$ are two set threshold for these two fuzzy matching functions.
\subsubsection{Evidence Summarization}
After obtaining the selected keywords set $\mathcal{S}_i$ corresponding to the $i$th piece of evidence $e_i$, our goal is to extract the information centered around these keywords within the evidence and discard the irrelevant. This extracted information is intended to be the most useful for verifying the claim. Essentially, extracting information centered on keywords is akin to uncovering the relationships between these keywords. Since meaningful relationships generally exist between multiple keywords, we focus our summaries on evidence containing at least two relevant keywords. Evidence with $|\mathcal{S}_i|<2$  will not be summarized, as we deem them unlikely to provide sufficient useful information. 
Still, we prompt the LLM to serve as the extractor and summarizer. Additionally, we equip the prompt with some examples to help the LLM better understand the compositional task and format its output. Formally, this process can be described as:
\begin{equation}
    a_i = \arg \max p(a_i|T_{ES},\mathcal{S}_i,e_i;\theta_{LLM}), |\mathcal{S}_i|\ge2
\end{equation}
where $a_i$ is the abstracted evidence from the raw evidence $e_i$, $\mathcal{S}_i$ is the selected keywords set. $T_{ES}$ is the prompt template used for Evidence Summarization:
\begin{mybox}
\textbf{Task Description}: Extract and summarize key information from sentences based on specified keywords. The output should be concise, directly related to the keywords, and devoid of extraneous details.\\
\textbf{Instructions}:
Carefully read the provided input sentence.
Use the specified keywords to guide your extraction of information.
Generate a summary that includes only the facts directly associated with the keywords.\\
\textbf{Examples}:\\
Input: Spam msubi is a popular snack and lunch food in Hawaii composed of a slice of grilled Spam on top of a block of rice, wrapped together with nori in the traditional of Japanese `omusubi'.\\
Keywords: spam, popular snack, lunch food, Hawaii.\\
Output: Spam is popular snack and lunch food in Hawaii.\\
\vspace{1.3mm}{{[}More Examples{]}}\\
Based on the following input, identify and list the key components as demonstrated in the examples.\\
\textbf{Input}: [Raw Evidence] ($e_i$)\\
\textbf{Keywords}: [Selected Keywords] ($\mathcal{S}_i$)\\
\textbf{Output}:
\end{mybox}

We perform the Keyword Selection and Evidence Summarization procedures for each piece of evidence, resulting in a set of abstracted evidence denoted as $\mathcal{A} =\{a_1,\cdots,a_{\sum_{i}{\{[}|\mathcal{S}_i|>2]}\}$. This set $\mathcal{A}$ is then combined with the raw evidence set $\mathcal{E}$ for subsequent verification tasks. The integration is essential because $\mathcal{A}$ encapsulates key elements crucial for directly validating the truthfulness of the claim, but may not contain all the information. By supplementing the raw evidence set, $\mathcal{A}$ can be considered an enhancement of the original evidence, providing a more concise and focused representation. Furthermore, it offers a shortcut for the LLM, reducing the complexity of subsequent inference processes.
\subsection{Claim Deconstruction}
Claim Deconstruction is the second component of \themodel. It takes the original claim as input and generates several subclaims that focus on different aspects. Simply asking an LLM to judge the claim's truthfulness is insufficient, as LLMs tend to compare the overall semantic meaning rather than scrutinize minor details. However, for claim verification, even the slightest error should result in the claim being judged as false, even if the semantic meaning remains largely unchanged. Relying solely on LLM judgments can only address obvious errors, failing to meet the objectives of comprehensive claim verification. By deconstructing the claim into subclaims, we can leverage LLMs to individually verify different aspects and details, thereby increasing the likelihood of identifying errors. Still, we prompt an LLM to deconstruct claim into subclaims:
\begin{equation}
    \{u_1,u_2,\cdots,u_r\} = \arg \max p(\mathcal{U}|T_{CD},c;\theta_{LLM})
\end{equation}
where $u_i$ means the $i$th subclaim from the potential subclaim set $\mathcal{U}$. $T_{CD}$ is the prompt template used for Claim Deconstruction:

\begin{mybox}
\textbf{Task Description}: Dissect a given claim into multiple atomic statements. These statements should be complete in meaning, devoid of uncertain pronouns, and retain all original details. Each atomic statement should stand alone and be independently verifiable.\\
\textbf{Examples}:\\
Claim: Spam is canned cooked meat by Hormel Foods Corporation is never used to make a popular snack and lunch food in Hawaii.\\
Output: \textbackslash{}n \#1 Spam is a canned cooked meat product manufactured by Hormel Foods Corporation. \textbackslash{}n \#2 Spam is not used to make a popular snack and lunch food in Hawaii.\\
\vspace{1.3mm}{{[}More Examples{]}}\\
Here is the claim given to you. Your answer should follow the format of above demonstrations. Each atomic statement should stand alone and be independently verifiable with as least pronouns as possible. Give your answer only, no explanation.\\
\textbf{Claim}: [Claim] ($c$)\\
\textbf{Output}:
\end{mybox}

\subsection{Subclaim Verification}\label{sec:SV}
The last step of \themodel is Subclaim Verification. After obtaining the set of subclaims $\{u_1,u_2,\cdots,u_r\}$, the truthfulness of the original claim can be verified by checking each subclaim individually. If any subclaim is false, the original claim is deemed false. The original claim is considered true only if all subclaims are correct. Mathematically, this can be represented as:
\begin{equation}
    \hat p = \begin{cases}
        \text{False} & \text{if }\exists i, f(u_i,\mathcal{A}\cup \mathcal{E})=\text{False}\\
        \text{True} & \text{Other}
    \end{cases}
\end{equation}
where $\hat p$ is the veracity prediction of the claim $c$, $\mathcal{A}$ and $\mathcal{E}$ are the abstracted evidence set and raw evidence set.
$f$ is the function used to verify the truthfulness of each subclaim $u_i$. We implement $f$ using LLM in a zero-shot manner, consistent with prior work~\cite{pan2023fact}. Mathematically, this can be written as:
\begin{equation}
    f(u_i,\mathcal{A}\cup \mathcal{E}) = \arg \max p(p_i|T_{SV},u_i,\mathcal{A}\cup \mathcal{E},c;\theta_{LLM})
\end{equation}
where $p_i=\text{True}\ or\ \text{False}$ represents the veracity prediction of the $i$th subclaim $u_i$,  and $T_{SV}$ is the prompt template used for Subclaim Verification:
\begin{mybox}
Given golden evidence: \\
{[}Abstracted Evidence \& Raw Evidence{]} ($\mathcal{A} \cup \mathcal{E}$)\\
\colorbox{gray!50}{In the saying of [Claim] ($c$)}.
Based on the golden evidence. Is it true that [Subclaim] ($u_i$)? (Yes or No)
\end{mybox}
The segment of the prompt highlighted in \colorbox{gray!50}{dark color} is optional. The decision to incorporate the context of the original claim for subclaim verification depends on the complexity of the claim. In our experiments, we observe that for complex claims, incorporating the original claim as context is more beneficial for verification. Further discussion on the optional prompt segment will be provided in the experimental section.
\section{Experiment}

\begin{table*}[t]
  \centering
    \begin{tabular}{c|c|ccc|c}
    \toprule[1.3pt]
    \multicolumn{2}{c|}{\textbf{Models}} & \textbf{HOVER-2}  & \textbf{HOVER-3}  & \textbf{HOVER-4}  & \textbf{FEVEROUS-S} \\
    \midrule[1.1pt]
    \multirow{5}[2]{*}{\parbox{3.5cm}{\centering \textbf{Pretrained/Fine-tuned Models}}} & BERT-FC & 53.40  & 50.90  & 50.86 & 74.71 \\
          & LisT5 & 56.15 & 53.76 & 51.67 & 77.88 \\
          & RoBERTa-NLI & \underline{74.62} & 62.23 & 57.98 & 88.28 \\
          & DeBERTaV3-NLI & \textbf{77.22} & 65.98 & 60.49 & \textbf{91.98} \\
          & MULTIVERS & 68.86 & 59.87 & 55.67 & 86.03 \\
    \midrule[1.1pt]
    \multirow{4}[2]{*}{\textbf{Vicuna}} & Zero-Shot & 64.08 & 64.63 & 59.59 & 81.69 \\
          & Few-Shot & 63.02 & 62.18 & 56.81 & 78.65 \\
          & ProgramFC & 66.07 & 60.35 & 56.74 & 87.51 \\
          \cmidrule{2-6}
          & + \themodel (our method) & 68.55\small\textcolor{green!58!black}{+4.47} & \underline{66.43}\small\textcolor{green!58!black}{+1.8} & \underline{63.42}\small\textcolor{green!58!black}{+3.83} & 89.37\small\textcolor{green!58!black}{+7.68} \\
    \midrule
    \multirow{4}[2]{*}{\textbf{Mixtral}} & Zero-Shot & 67.86 & 64.03 & 62.09 & 85.06 \\
          & Few-Shot & 66.59 & 63.59 & 62.55 & 88.49 \\
          & ProgramFC & 59.97 & 61.75 & 59.82 & 81.76 \\
          \cmidrule{2-6}
          & + \themodel (our method) & 73.17\small\textcolor{green!58!black}{+5.31} & \textbf{69.40}\small\textcolor{green!58!black}{+5.37} & \textbf{67.78}\small\textcolor{green!58!black}{+5.69} & \underline{89.52}\small\textcolor{green!58!black}{+4.46} \\


    \bottomrule[1.06pt]
    \end{tabular}%
    \caption{Comparison of baseline models on subsets of HOVER dataset and FEVEROUS-S dataset in terms of Macro-F1 score. HOVER-2 represents the 2-hops subset of the HOVER dataset. Green numbers show improvement over zero-shot performance when our method is applied to backbone LLMs, as the verification process of \themodel is also zero-shot.}
  \label{tab:main}%
\end{table*}%

\subsection{Dataset}
In line with existing research, we have selected two publicly available datasets to assess the performance of \themodel. Evaluation is carried out using the validation set. The chosen datasets are HOVER~\cite{jiang2020hover} and FEVEROUS-S~\cite{aly2021feverous}. 
\begin{itemize}
    \item \textbf{HOVER} \xspace The HOVER dataset comprises claims that necessitate verification through multiple pieces of evidence and multi-hop reasoning. It is organized into three subsets, each corresponding to a different level of reasoning complexity based on the number of hops. Specifically, the two-hop subset (HOVER-2) consists of 1,126 claims, the three-hop subset (HOVER-3) comprises 1,835 claims, and the four-hop (HOVER-4) subset includes 1,039 claims.
    
    \item \textbf{FEVEROUS-S} \xspace FEVEROUS is a fact-checking dataset designed to validate claims using both structured and unstructured data sources. Our experimentation is focused on a subset of FEVEROUS, known as FEVEROUS-S, which exclusively involves claims that rely on unstructured data. In terms of claim complexity, it is noted that the claims in the HOVER dataset exhibit higher complexity compared to those in FEVEROUS-S.
\end{itemize}
Given our emphasis on the claim verification task, all experiments are executed using the evidence provided within the dataset (referred to as golden evidence). The performance is assessed using the Macro-F1 score as the evaluation metric.  

\subsection{Baselines}
\themodel is a versatile framework that can be adapted to various existing large language models. In order to ensure credibility and inclusivity, we have selected two open-source LLMs with differing parameter sizes as the foundational backbone for \themodel. These models are \textbf{Vicuna-13B}~\cite{vicuna2023} and \textbf{Mixtral-8x7B}~\cite{jiang2024mixtral}. Our experimentation includes zero-shot and few-shot trials using these backbone models. Furthermore, we conduct experiments with these two language models within another framework, ProgramFC~\cite{pan2023fact}, which prompts LLMs to generate and execute programs for the purpose of claim verification.

The following pretrained or fine-tuned models are also considered as baseline models:
\begin{itemize}
    \item BERT-FC~\cite{soleimani2020bert}: Pretrained BERT model~\cite{devlin2018bert} tailored for fact-checking tasks.
    \item LisT5~\cite{jiang2021exploring}: Pretrained T5 model~\cite{raffel2020exploring} specialized for fact-checking tasks.
    \item RoBERTa-NLI~\cite{nie2020adversarial}: Pretrained RoBERTa-large model~\cite{liu2019roberta} fine-tuned on four natural language inference datasets.
    \item DeBERTaV3-NLI~\cite{he2021debertav3}: Pretrained DeBERTaV3 model fine-tuned on FEVER~\cite{thorne2018fever} and four natural language inference datasets.
    \item MULTIVERS~\cite{wadden2022multivers}: A LongFormer model~\cite{beltagy2020longformer} fine-tuned on the FEVER dataset.
\end{itemize}


\subsection{Implementation Details}
In the Keyword Selection process, both similarity score thresholds ($t_1$ and $t_2$) are set to $60$ (maximum is 100) to ensure the retention of important keywords.

\themodel conducts the verification process in a zero-shot manner but includes in-context examples in the prompts for Evidence Abstraction and Claim Deconstruction. To ensure fair experimentation, the model does not use any examples that are not utilized by the baseline models. Few-shot experiments with backbone models use the same examples as prompts in ProgramFC. Examples for Evidence Abstraction and Claim Deconstruction are rephrased from ProgramFC to suit task requirements. Subclaim Verification uses the optional prompt component $T_{SV}$ for the HOVER dataset but not for the FEVEROUS-S dataset. A more detailed discussion on $T_{SV}$ is provided in Section \ref{sec:claim}.

Since open-source models are utilized, all experiments are conducted on a local machine server equipped with an AMD EPYC 7742 (256) @ 2.250GHz CPU and NVIDIA RTX 3090 (24G) GPUs (Vicuna-13B experiments require two GPUs, while Mixtral-8x7B necessitates a minimum of five GPUs). A temperature of 0.05 is utilized to reduce randomness, while all other hyperparameters in sampling output of LLMs remain default.

\begin{table}[tb]
  \centering
    \setlength{\tabcolsep}{3pt}
    \begin{tabular}{ccc|ccc}
    \toprule[1.3pt]
    $\textbf{CD}$    & $\textbf{EA}_v$ & $\textbf{EA}_m$ & \textbf{HOVER-2} & \textbf{HOVER-3} & \textbf{HOVER-4} \\
    \midrule[1.1pt]
    $\times$     & $\times$     & $\times$     & 64.08 & 64.63 & 59.59 \\
    \midrule
    \Checkmark     & $\times$     & $\times$     & 66.90  & 65.61 & 62.46 \\
    $\times$     & \Checkmark     & $\times$     & 64.25 & 64.98 & 61.52 \\
    $\times$     & $\times$     & \Checkmark     & 63.99 & 65.15 & 63.90 \\
    \Checkmark     & \Checkmark     & $\times$     & \textbf{68.55} & 66.43 & 63.42 \\
    \Checkmark     & $\times$     & \Checkmark     & 66.25 & \textbf{66.97} & \textbf{64.23} \\
    \bottomrule[1.06pt]
    \end{tabular}%
  \caption{Ablation study on \themodel using Vicuna as the subclaim verifier. $\text{CD}$ refers to Claim Deconstruction with Vicuna, while $\text{EA}_v$/$\text{EA}_m$ denotes Evidence Abstraction with Vicuna/Mixtral. Macro-F1 scores are reported.}
  \label{tab:ablation}%
\end{table}%

\subsection{Overall Performance}
Table \ref{tab:main} presents the results of our method and various baseline models. The data clearly demonstrates that \themodel consistently and substantially improve model performance across both datasets, using either Vicuna or Mixtral as the backbone model.

Compared to pretrained/fine-tuned models, zero-shot LLMs do not exhibit advantage, especially compared to DeBERTaV3. However, applying our proposed model, \themodel, to these LLMs demonstrates a more pronounced advantage in complex tasks.
On simpler datasets like FEVEROUS-S and HOVER-2 (2-hop reasoning), \themodel-equipped models perform comparably to pretrained/fine-tuned models. But for more complex tasks like HOVER-3 and HOVER-4, \themodel-equipped LLMs show a distinct advantage.

In comparison to other LLM-based approaches for claim verification, our model demonstrates superior stability. The few-shot technique does not consistently improve performance, aligning with prior research~\cite{hu2023large}. ProgramFC's reliance on LLMs' program generation and execution capabilities makes it less adaptable and more sensitive to intermediate errors compared to the model.

\subsection{Ablation Study}

\subsubsection{The Impact of Evidence Abstraction and Claim Deconstruction}
To navigate the ``noisy'' crowd of evidence and claim, \themodel contains two key components: Evidence Abstraction and Claim Deconstruction. In this section, we conduct ablation studies to understand the contribution of each component. Removing the Evidence Abstraction component eliminates the use of the abstracted evidence set $\mathcal{A}$ in verification, while removing Claim Deconstruction results in direct assessment of the claim's truthfulness without generating subclaims. We show the results of these ablation experiments with Vicuna as the subclaim verifier in Table \ref{tab:ablation}.

The results indicate that utilizing either the Evidence Abstraction or Claim Deconstruction component independently leads to improvements in the backbone LLM's performance in claim verification. Combining both components further enhances the model's performance.
Furthermore, we observe that the choice of the LLM used for the Evidence Abstraction component also affects the model's performance. Specifically, using Mixtral for Evidence Abstraction enhances the large language model's ability to evaluate complex claims more significantly than using the Vicuna model. 

\begin{table}[t]
  \centering
    \begin{tabular}{c|ccc}
    \toprule[1.3pt]
    \textbf{Model} & \textbf{HOVER-2} & \textbf{HOVER-3} & \textbf{HOVER-4 }\\
    \midrule[1.1pt]
    \textbf{Full Model}  & \textbf{68.55} & \textbf{66.43} & \textbf{63.42} \\
    \midrule
    w/o Keyword & 64.62 & 63.97 & 60.48 \\
    w/o Selection & 63.74 & 63.73 & 62.78 \\
    w/o Raw &65.80 & 65.56 & 60.17\\
    \bottomrule[1.06pt]
    \end{tabular}%
    \caption{Macro-F1 scores of different Evidence Abstraction settings using Vicuna as the backbone model. ``w/o Keyword'' indicates abstraction without keyword guidance, relying solely on the claim. ``w/o Selection'' indicates abstraction using all keywords without the Keyword Selection process. ``w/o Raw'' indicates using solely the abstracted evidence set $\mathcal{A}$ for verification without the raw evidence set $\mathcal{E}$.}
  \label{tab:Key}%
\end{table}%

\subsubsection{The Rationale Behind Keyword Selection}\label{section:WK}
We employ a keyword-based method in Evidence Abstraction. As elucidated in Section \ref{section:EA} and Figure \ref{fig:explain}, extracting keywords from the claim to guide evidence abstraction serves to preempt potential conflicts between claim and evidence content. Selecting relevant keywords by fuzzy matching reduces LLMs' tendency to generate content not supported by the evidence. To further assess this methodology, we examine Evidence Abstraction performed without keyword guidance (w/o Keyword) and without Keyword Selection (w/o Selection). In the w/o Keyword scenario, the LLM summarizes raw evidence based solely on the claim (Part I of Figure \ref{fig:explain}). In the w/o Selection scenario, the LLM uses all keywords for guidance (Part II of Figure \ref{fig:explain}). Results are presented in Table \ref{tab:Key}.

As shown in the table, the performance of \themodel significantly deteriorates when either keyword guidance or keyword selection is omitted. This highlights the crucial role of selecting keywords as guidance in enhancing the effectiveness of Evidence Abstraction.

\begin{table}[tb]
\setlength{\tabcolsep}{2.3pt}
  \centering
  \resizebox{\columnwidth}{!}{
    \begin{tabular}{c|ccc|c}
    \toprule[1.3pt]
   \textbf{Model} & \textbf{HOVER-2}  & \textbf{HOVER-3} & \textbf{HOVER-4} & \textbf{FS-S} \\
    \midrule[1.1pt]
    \themodel w/ Claim & \textbf{68.55} & \textbf{66.43} & \textbf{63.42} & 82.53 \\
    \themodel w/o Claim& 68.37 & 62.57 & 55.7  & \textbf{89.37} \\
    \bottomrule[1.06pt]
    \end{tabular}%
   }
    \caption{Macro-F1 scores of different Subclaim Verification settings using Vicuna as the backbone model. ``w/ Claim'' indicates the use of the optional part in the prompt $T_{SV}$, which mentions the original claim. ``w/o Claim'' indicates its absence. FS-S refers to the FEVEROUS-S dataset.}
  \label{tab:claim}%
\end{table}%

\subsubsection{Effectiveness of Concatenating Raw Evidence Set $\mathcal{E}$}  
In Evidence Summarization, the abstracted evidence set $\mathcal{A}$ is considered an augmentation to the raw evidence set. Analyzing Table \ref{tab:Key}, it is apparent that using only the abstracted evidence set $\mathcal{A}$ (w/o Raw) results in suboptimal performance compared to using the combined set $\mathcal{A} \cup \mathcal{E}$ (Full Model). This discrepancy arises because the keyword-based abstraction method, while capturing crucial information, may overlook hard-to-identify details. Therefore, the strategy of concatenating $\mathcal{E}$ and $\mathcal{A}$ proves to be an effective approach.

\subsubsection{Analysis of Optional Claim Context in Subclaim Verification}\label{sec:claim}
In Section \ref{sec:SV}, we mentioned that the prompt used in the Subclaim Verification process includes an optional segment: ``\textit{In the saying of [Claim] ($c$)}''. In our \themodel experiments, we included this optional component for the HOVER dataset but not for the FEVEROUS-S dataset. The presence of this optional component significantly impacts the Subclaim Verification step.

 As depicted in Table \ref{tab:claim}, its inclusion enhances model performance in complex reasoning scenarios such as HOVER-3 and HOVER-4, while showing minimal improvement in simpler datasets like FEVEROUS-S. Complex datasets may feature intricate logical relationships in claims, where nested logic, like ``\textit{The coach, who worked with the Seattle Seahawks, was an employee of the Cleveland Browns},'' could lead to LLM deconstructing a subclaim as ``\textit{The coach was an employee of the Cleveland Browns.}'' In such cases, providing comprehensive context is crucial. In essence, for straightforward scenarios, minimizing additional contextual information optimally leverages the LLM's reasoning abilities in subclaim verification. Conversely, in complex scenarios, offering extensive context proves more effective, aligning with common-sense judgment.

\section{Conclusion}
In this paper, we introduce the \themodel framework to enhance LLMs in claim verification task. We address the challenge posed by ``noisy crowd'' of evidence and claims that can negatively impact LLMs' performance. To address this, we propose Evidence Abstraction to extract essential information from noisy evidence and Claim Deconstruction to verify distinct aspects of the original claim individually. We present an abstraction method based on selected keywords to mitigate conflicts between claims and evidence, reducing the risk of generating unsupported content during evidence abstraction. We also examine the impact of incorporating the original claim into the subclaim verification process. Our validation on two datasets using two open-source LLMs shows the effectiveness of the \themodel framework.
\bibliography{main}

\end{document}